\documentclass[final]{cvpr}

\usepackage{times}
\usepackage{epsfig}
\usepackage{graphicx}
\usepackage{amsmath}
\usepackage{amssymb}
\usepackage{subfigure}
\usepackage{capt-of}
\usepackage[ruled]{algorithm2e}

\SetAlFnt{\small}
\SetAlCapFnt{\small}
\SetAlCapNameFnt{\small}
\SetAlCapHSkip{0pt}

\usepackage{amsmath,amssymb,amsfonts,dsfont,bm,mathrsfs}
\usepackage{booktabs,multirow,adjustbox}

\newcommand{\w}{{\rm\bf w}}       % notation of disentangled latent code in StyleGAN.
\newcommand{\F}{{\rm\bf F}}       % notation of the generated features.
\newcommand{\m}{{\rm\bf m}}       % notation of the segmentation masks.
\usepackage[pagebackref=true,breaklinks=true,colorlinks,bookmarks=false]{hyperref}

\def\cvprPaperID{11} % *** Enter the CVPR Paper ID here
\def\confYear{CVPR 2021}

\begin{document}

%%%%%%%%% TITLE
\title{Decorating Your Own Bedroom: Locally Controlling Image Generation \\ with Generative Adversarial Networks}

\author{
  Chen Zhang$^1$ \quad Yinghao Xu$^2$ \quad Yujun Shen$^2$ \\
  $^1$Zhejiang University \quad $^2$The Chinese University of Hong Kong \\
  {\tt\small daisy\_chen@zju.edu.cn \quad \{xy119, sy116\}@ie.cuhk.edu.hk}
}

%%%% Figure: Teaser
\twocolumn[{
\renewcommand\twocolumn[1][]{#1}
\vspace{-10pt}
\maketitle
\vspace{-25pt}
\begin{center}
    \includegraphics[width=0.95\linewidth]{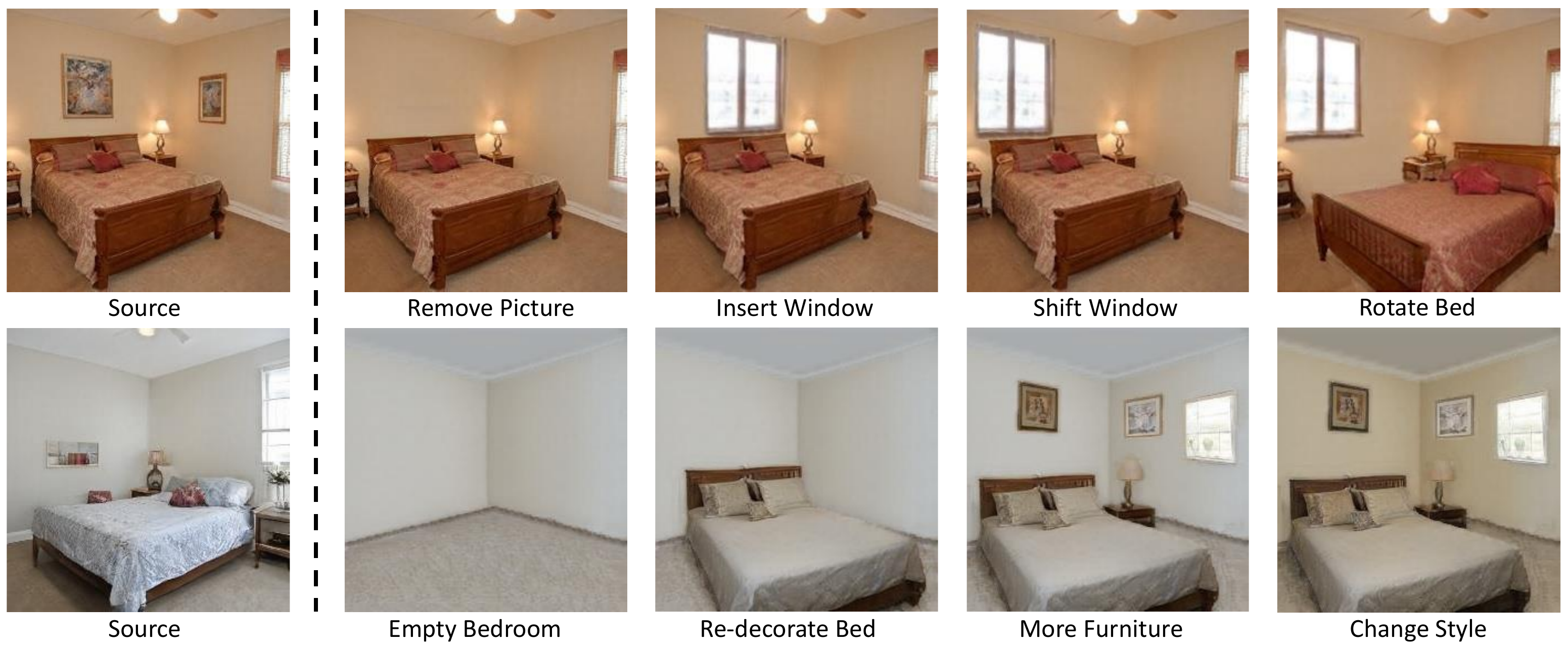}
    \captionof{figure}{
        \textbf{Local editing} achieved by LoGAN using the well-trained StyleGAN2~\cite{stylegan2} model for bedroom synthesis.
        Top row presents some basic operations on individual objects, while bottom row demonstrates clearing out the room, refurnishing, and further changing style.
    }
    \label{fig:teaser}
\end{center}
}]

%%%%%%%%% ABSTRACT
\begin{abstract}
\vspace{-10pt}
Generative Adversarial Networks (GANs) have made great success in synthesizing high-quality images.
However, how to steer the generation process of a well-trained GAN model and customize the output image is much less explored.
It has been recently found that modulating the input latent code used in GANs can reasonably alter some variation factors in the output image, but such manipulation usually presents to change the entire image as a whole.
In this work, we propose an effective approach, termed as LoGAN, to support local editing of the output image.
Concretely, we introduce two operators, \textit{i.e.}, content modulation and style modulation, together with a priority mask to facilitate the precise control of the intermediate generative features.
Taking bedroom synthesis as an instance, we are able to seamlessly remove, insert, shift, and rotate the individual objects inside a room.
Furthermore, our method can completely clear out a room and then refurnish it with customized furniture and styles.
Experimental results show the great potentials of steering the image generation of pre-trained GANs for versatile image editing.
\end{abstract}

\begin{figure*}[t]
    \centering
    \includegraphics[width=1\linewidth]{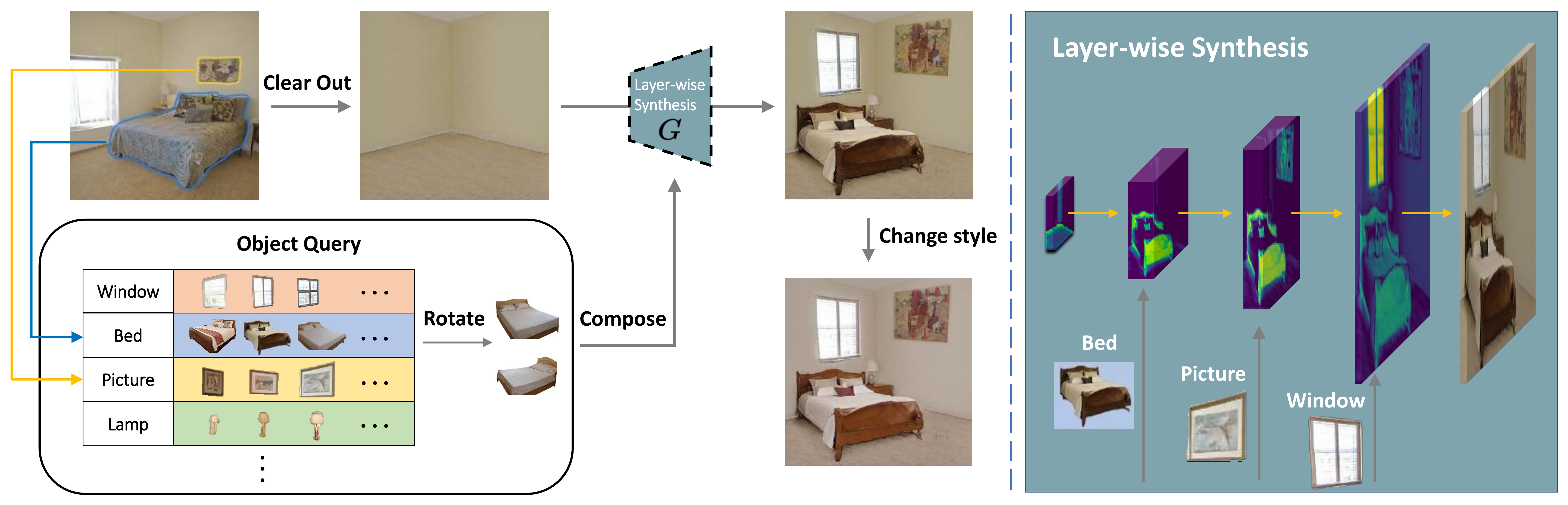}
    \caption{
        \textbf{Overview of the LoGAN pipeline} to re-decorate a bedroom.
        It consists of four steps: (i) clearing out the room based on layout prediction, (ii) selecting individual objects from the candidate pool, (iii) composing the bedroom by inserting each object to the target position at the proper layer, and (iv) rendering the recomposed scene with different styles.
    }
    \label{fig::pipeline}
\vspace{-10pt}
\end{figure*}

%%%%%%%%% BODY TEXT
\section{Introduction}\label{sec:introduction}
Recent years have witnessed the significant advance of Generative Adversarial Networks (GANs) in image synthesis~\cite{pggan,stylegan,biggan,stylegan2}.
However, precisely steering the generation with a pre-trained GAN model and customizing the output image remain unsolved.
Recent studies focus on global editing but adjusting fragments is more critical in many practical cases.
In this paper, we propose an approach for the \textit{local} control of the image generation with GANs, termed as \textit{LoGAN}.
Different from prior work~\cite{bau2021paint, bau2020semantic}, LoGAN performs manipulation from the feature space instead of the initial latent space.
We find that the intermediate feature maps developed by GAN generators effectively encode rich spatial information, which supports editing the output within local regions.
With the state-of-the-art StyleGAN2~\cite{stylegan2} model learned for bedroom synthesis, we design content and style modulation to implement flexible object composition. 
In addition, we introduce the priority mask to handle the spatial relationship between different objects.
In this way, we can impressively edit the single object and further clear out then re-decorate the whole bedroom with customized furniture and styles, as shown in Fig.~\ref{fig:teaser}.
%

%%%% Section: Related Work
\vspace{-2pt}
\subsection{Related Work}\label{sed:related-work}
\vspace{-2pt}
%%%%
\noindent\textbf{Generative Adversarial Networks.}
GAN~\cite{gan} is proposed to learn the mapping from a pre-defined latent space to the observed image space via adversarial training.
Many GAN variants have been proposed to improve the synthesis quality and training stability~\cite{wgan,wgan_gp,began,sagan,sngan,pggan,stylegan,biggan,stylegan2}.
However, controlling the generation process of well-trained GANs is much less explored.

\vspace{2pt}
\noindent\textbf{Image Editing with GANs.}
It has been recently found that GANs spontaneously learn rich semantics when trained to synthesize images~\cite{interfacegan,ganalyze,steerability,higan}.
These semantics are commonly characterized as some particular directions in the latent space.
But manipulation in the latent space~\cite{interfacegan, higan} tends to change the whole image.
Differently, some recent studies~\cite{suzuki2018spatially,collins2020editing,abdal2020image2stylegan++} propose to utilize the spatial feature maps generated by GANs for local editing.
Nevertheless, they mainly focus on transferring one object to another~\cite{suzuki2018spatially,abdal2020image2stylegan++} or stylizing one object using another as the reference~\cite{collins2020editing}.
Some work~\cite{tsai2017deep,lin2018st,zhan2019spatial,nguyen2020blockgan, tseng2020retrievegan} particularly explores image compositing which fuses some unrelated fragments together into a reasonable image.
Some studies~\cite{zhang2018generative,lee2020maskgan,jo2019sc} focus on locally editing face images, like changing the nose of or adding bangs to a target face.
In comparison, bedroom synthesis studied in this work is more challenging due to the high complexity of the room layout and the large diversity of the objects inside.
More importantly, LoGAN supports customizing the room decoration without any retraining.

%%%% Section: Method
\vspace{-2pt}
\section{Framework of LoGAN}\label{sec::method}
\vspace{-2pt}
%%%%
% In this section, we introduce the proposed LoGAN. 
%
The framework is shown as Fig.~\ref{fig::pipeline}.
Precisely, with a content modulation operator $\mathtt{CMod}(\cdot, \cdot)$, a style modulation operator $\mathtt{SMod}(\cdot, \cdot)$, and an effective priority mask, LoGAN achieves local control of the image generation with GANs.

\begin{figure}[t]
    \centering
    \includegraphics[width=1\linewidth]{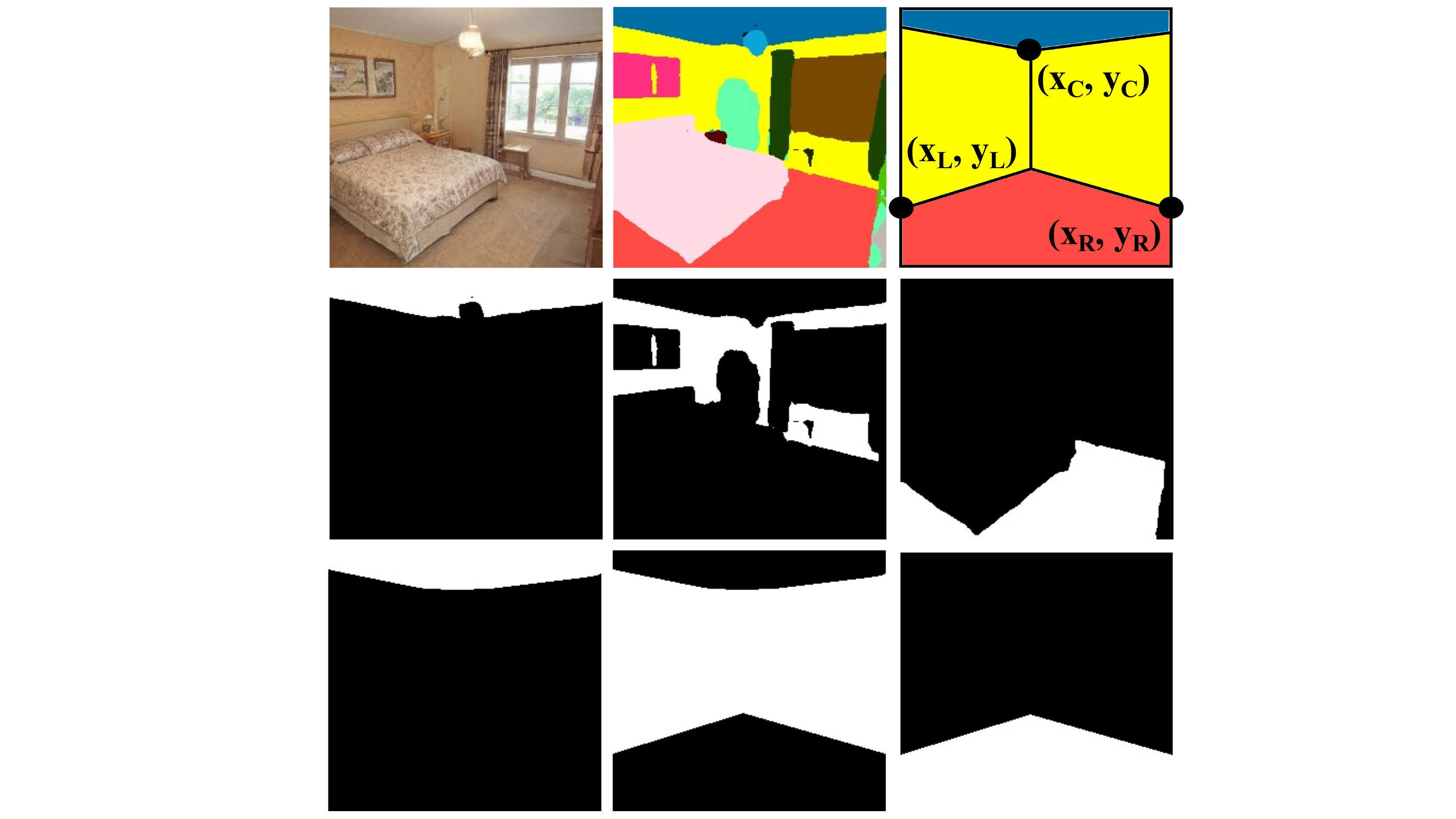}
    \caption{
        \textbf{Diagram of parsing the bedroom layout} with an image segmentation model~\cite{deeplabv3}.
        Top row shows the segmentation map for backgrounds while the bottom row visualizes the parsed layout.
    }
    \label{fig:basic_layout}
\vspace{-8pt}
\end{figure}

\vspace{2pt}
\noindent\textbf{Preliminary on Style-Based Generator.}\label{sec::pre}
The style-based generator~\cite{stylegan,stylegan2} proposes to adjust the layer-wise image styles using Adaptive Instance Normalization (AdaIN)~\cite{adain} operation.
Accordingly, the layer-wise style control can be formulated as 
\vspace{-6pt}
\begin{align}
    \label{eq::style}
    \mathtt{A}(\F^{(\ell)}, \w^{(\ell)}) = \
    \{\mathtt{AdaIN}(\F^{(\ell)}_c,T^{(\ell)}(\w^{(\ell)}))\}_{c=1}^C,
\vspace{-6pt}
\end{align}
where $c$ and $C$ respectively denote the channel index and total number of feature channels, and $T(\cdot)$ means the learned affine transformations who specialize the latent code into styles.
In this way, given the spatial feature maps $\F^{(\ell)}$ for the particular $\ell$-th layer, we can adjust the overall style by altering the corresponding latent code $\w^{(\ell)}$.

\vspace{2pt}
\noindent\textbf{Layout Parsing.}
In order to predict the layout of bedroom, we propose a background completion approach by solving the occlusion problem.
As shown in Fig.~\ref{fig:basic_layout}, we treat ceilings, walls, and floors as the basic components of background.
Based on the segmented map from semantic segmentation model~\cite{deeplabv3}, we first find the convex closure for ceiling, then get the key point $(x_C, y_C)$.
In the meantime, we find $(x_L, y_L)$ and $(x_R, y_R)$ from both sides of the image and form the floor boundary with assigned slope.
Such auto-searching method is capable of getting well-shaped bedroom layouts, which are very close to the real cases.

\vspace{2pt}
\noindent\textbf{Content Modulation.}
To enable content editing, we propose a content modulation operator based on the generative features as
\vspace{-8pt}
\begin{equation}
    \label{eq::content_mod}
    \mathtt{CMod}(\F^{(\ell)}\!,\! (\F_o^{(\ell)}\!,\! \m_o^{(\ell)})\!) \!=\! \F^{(\ell)} \!\odot\! (1 \!-\! \m_o^{(\ell)}\!) \!+\!
                                                                 \F_o^{(\ell)} \!\odot\! \m_o^{(\ell)},
\vspace{-8pt}
\end{equation}
where $\m_o^{(\ell)}$ and $\F_o^{(\ell)}$ represent the mask and feature map of the editing region.
The $\odot$ denotes the pixel-wise multiplication along the channel index.
With Eq.~\eqref{eq::content_mod}, we can achieve object insertion and removal by replacing $\F_o^{(\ell)}$ with $\F_b^{(\ell)}$ to refill the erased region with background feature.

\vspace{2pt}
\noindent\textbf{Object Clustering and Rotation.}
To study object generating, we firstly vectorize the object region with downsampled height and width as $(H_s, W_s)$.
Then we perform k-means algorithm to find $M$ clusters representing different shapes of the object.
Finally we propose to interpolate the layer-wise latent code $\w^{(\ell)}$ between clustering centers as
\vspace{-8pt}
\begin{equation}
    \label{eq::rotate_code}
    \w_s^{(\ell)}=\w_l^{(\ell)}+\frac{s}{S}(\w_r^{(\ell)}-\w_l^{(\ell)}), s=0,1,\cdots,S,
\vspace{-8pt}
\end{equation}
where $\w_l^{(\ell)}$ and $\w_r^{(\ell)}$ respectively denote the latent codes of clustered centers on left and right object poses.
Taking $\w_s^{(\ell)}$ as the input, we manage to rotate the object inside an image.

\vspace{2pt}
\noindent\textbf{Priority Mask.}
Real cases show the object regions often have overlaps with each other, so we introduce ``priority'' into the mask called priority mask $\m_o^{(\ell)}(p)$.
We simply design the priority of different objects according to their decoration sequence in common sense.
Larger value of $p$ determines the higher priority of execution order in the modulation procedure.
The priority mask is computed as
\vspace{-8pt}
\begin{equation}
    \m_o^{(\ell)}=\m_o^{(\ell)}(p) \times (1-\sum_{o'\neq o}\m_{o'}^{(\ell)}(p')\mathds{1}_{p'>p})^{+},
    \label{eq::priority_mask}
\vspace{-8pt}
\end{equation}
where $\mathds{1}$ denotes the indicator function.
Such design turns object masks $\{\m_o^{(\ell)}\}_{o=1}^N$ into disjoint regions, which ensures the customization of the local synthesis control.

\vspace{2pt}
\noindent\textbf{Style Modulation.}
We should consider to locally control the style when editing an object.
To this end, we propose the style modulation operator $\mathtt{SMod}(\cdot, \cdot)$, formulated as
\vspace{-6pt}
\begin{align}
    \label{eq::style_mod}
     & \mathtt{SMod}((\F^{(\ell)}, \w^{(\ell)}), (\F_o^{(\ell)}, \w_o^{(\ell)}, \m_o^{(\ell)})) \nonumber \\
  =\ &\mathtt{A}(\F^{(\ell)}, \w^{(\ell)}) \odot (1 \!-\! \m_o^{(\ell)}) \!+\! \mathtt{A}(\F_o^{(\ell)},\w_o^{(\ell)}) \odot \m_o^{(\ell)}.
\vspace{-8pt}
\end{align}
Here, $\w^{(\ell)}$ is still used to control the global style, same as in Eq.~\eqref{eq::style}.
Differently, LoGAN assigns a code $\w_o^{(\ell)}$ for each particular region $\m_o^{(\ell)}$ and hence supports controlling the style of individual objects independently.

\vspace{2pt}
\noindent\textbf{Layer-wise Synthesis.}
At length, our overall local editing system embeds $\mathtt{CMod}(\cdot, \cdot)$ and $\mathtt{SMod}(\cdot, \cdot)$ into the layer-wise generator $G(\cdot) \triangleq G^{(L)} \circ G^{(L-1)} \circ \cdots \circ G^{(1)}(\cdot)$.
The layer-wise generation process is summarized in Algorithm~\ref{alg:synthesis}.
\begin{algorithm}[!ht]  
  \caption{Local editing with layer-wise synthesis.}  
  \label{alg:synthesis}  
    \KwIn{initial feature map: $\F^{(1)}$, \\
        \hspace{24pt}layer-wise base latent code: $\w^{(*)}$, \\
        \hspace{24pt}layer-wise object list: $\text{L}[*]$, \\
        \hspace{24pt}layer-wise object: $\{\F_o^{(*)}\}_o, \{\w_o^{(*)}\}_o, \{\m_o^{(*)}\}_o$.
    }
    \KwOut{final synthesis: $\F^{(L+1)}$.}
    \For {$\ell$-th layer in generator} {
        \For {object $o$ in $\text{L}[\ell]$ with $(\F_o^{(\ell)}, \w_o^{(\ell)}, \m_o^{(\ell)})$} {
            update priority mask: $\m_o^{(\ell)} \gets \m_o^{(\ell)}$; \\
            edit content: $\F^{(\ell)} \gets \mathtt{CMod}(\F^{(\ell)}, (\F_o^{(\ell)}, \m_o^{(\ell)}))$; \\
            edit style: $\F^{(\ell)} \gets \mathtt{SMod}((\F^{(\ell)}, \w^{(\ell)}), (\F_o^{(\ell)}, \w_o^{(\ell)}, \m_o^{(\ell)}))$;
        }
        generate the next feature map $\F^{(\ell+1)} \gets G^{(\ell)}(\F^{(\ell)})$;
    }
\end{algorithm}

%%%% Section: Experiments
\vspace{-2pt}
\section{Experiments}\label{sec:experiments}
\vspace{-2pt}
%%%%
Here we validate the effectiveness of LoGAN on the StyleGAN2~\cite{stylegan2} model trained on LSUN~\cite{lsun} bedrooms.

\definecolor{amber}{rgb}{1.0, 0.75, 0.0}
\begin{figure}[t]
\vspace{-10pt}
    \centering
    \includegraphics[width=1.0\linewidth]{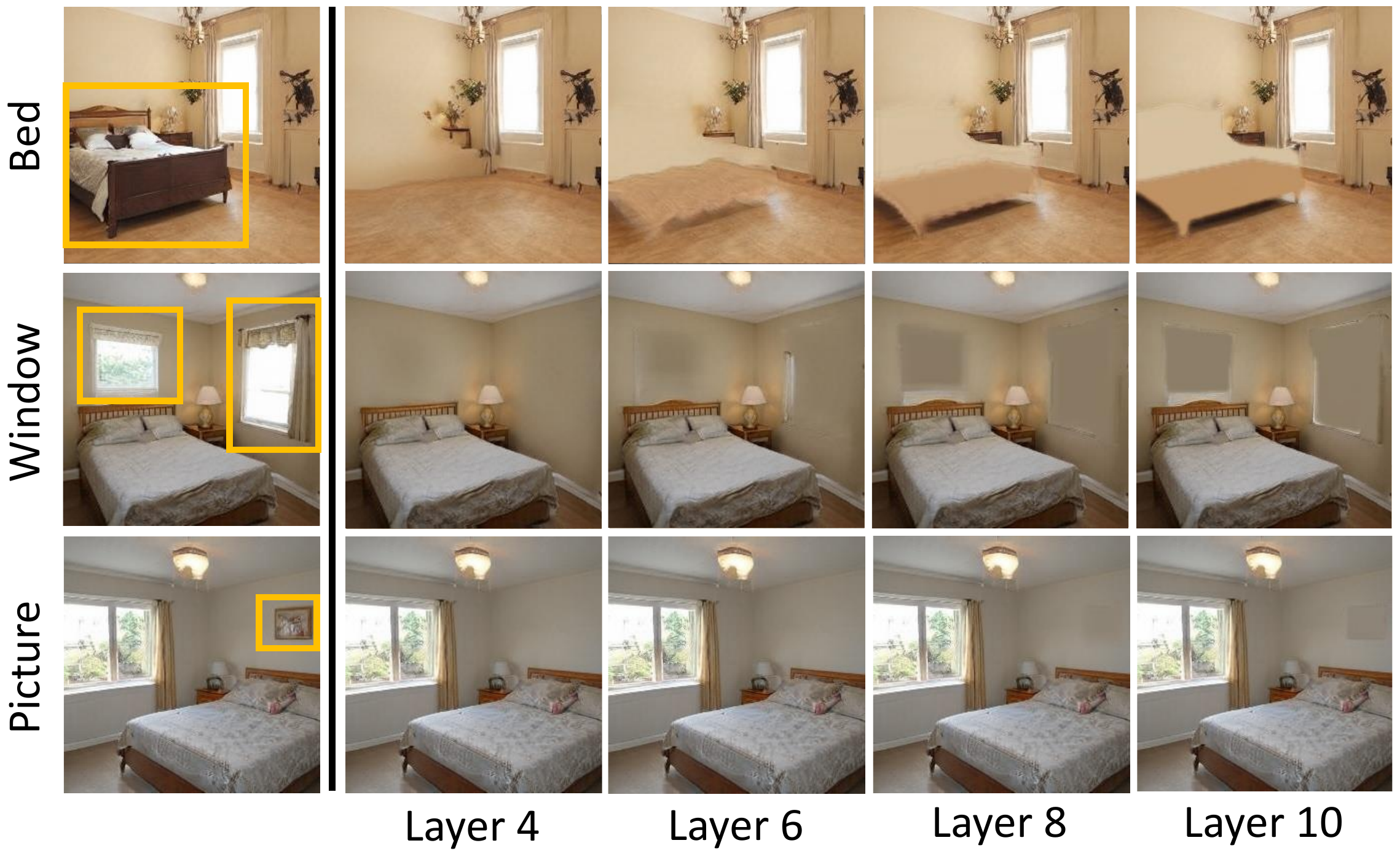}
    \caption{
        \textbf{Layer-wise analysis on object removal}.
        We remove the objects highlighted by the \textbf{\textcolor{amber}{orange}} boxes from the source images in the first column.
        The remaining columns show the results of removing a particular object at different layers.
    }
    \label{fig::single_erase}
    \vspace{-6pt}
\end{figure}

\vspace{-2pt}
\subsection{Single Object Removal}
\vspace{-2pt}
Fig.~\ref{fig::single_erase} shows the results of removing a bed, window, and picture from the source image.
For each object category, we perform the removal operation at the 4-th, 6-th, 8-th, and 10-th layers respectively.
We observe that manipulation at 4-th layer could better blend background feature with object feature, leading to more satisfying results.
That is because the convolutional kernels at early layers have larger receptive field.
By contrast, editing at later layers makes the removal region much sharper, especially at the mask boundaries.
As a result, we recommend removing objects at the 4-th layer.

\vspace{-2pt}
\subsection{Creating an Empty Bedroom}
\vspace{-2pt}

\begin{figure}[t]
    \centering
    \includegraphics[width=1.0\linewidth]{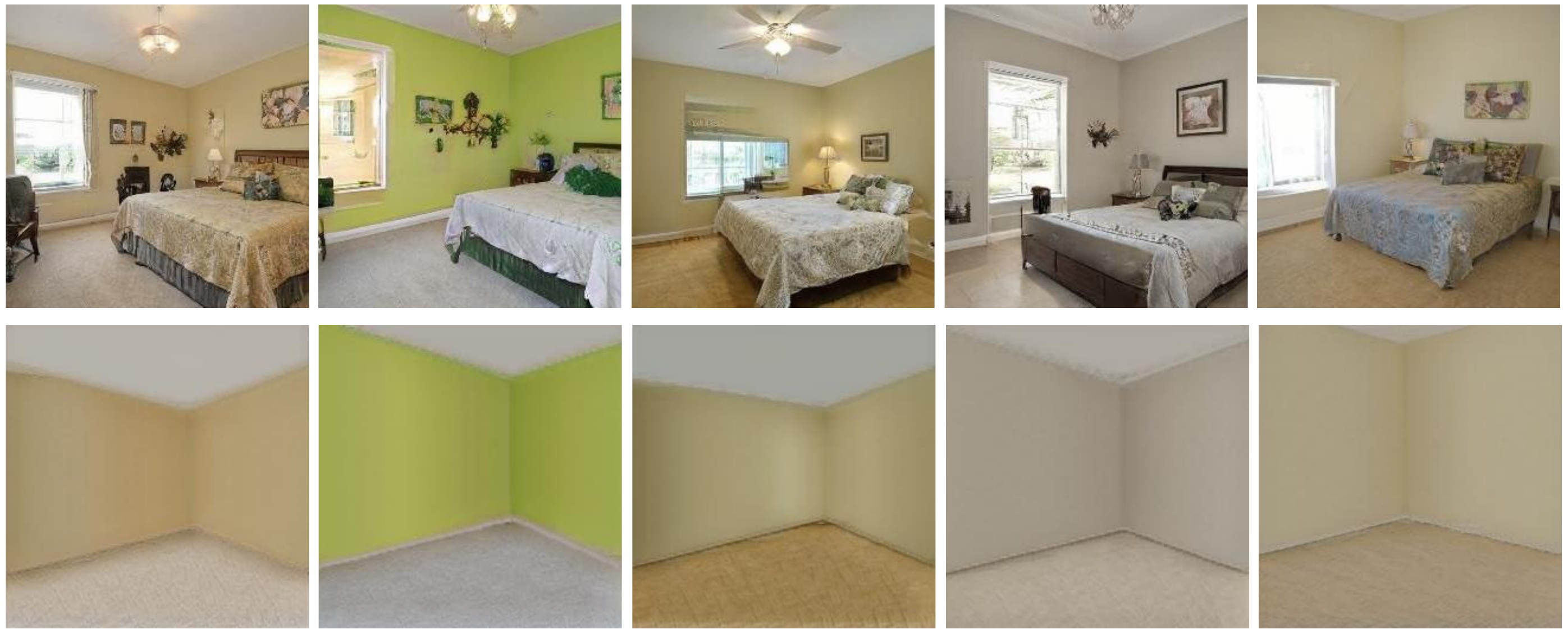}
    \caption{
        \textbf{Empty bedrooms} created by LoGAN, where we remove all objects inside the bedroom based on the parsed layout.
    }
    \label{fig::empty_bedroom}
    \vspace{-6pt}
\end{figure}

As illustrated in Fig.~\ref{fig::empty_bedroom}, LoGAN creates some empty bedrooms for example.
It's noticed that the layout of the empty room satisfyingly matches the source room and the texture of the background is also well preserved from the source image, demonstrating the effectiveness of the proposed layout parsing algorithm and the feature manipulation algorithm.

\vspace{-2pt}
\subsection{Clustering and Rotating Beds}
\vspace{-2pt}

The results evaluate that clustering on downsampled object mask could easily discriminate the shape and orientation of bed.
Besides, we manage to rotate the bed in source image to other poses while preserving the original color and texture.
We conclude that interpolating the latent codes between clustering centers is able to control the object then help to realize the rotation.

\begin{figure}[t]
    \centering
    \includegraphics[width=1.0\linewidth]{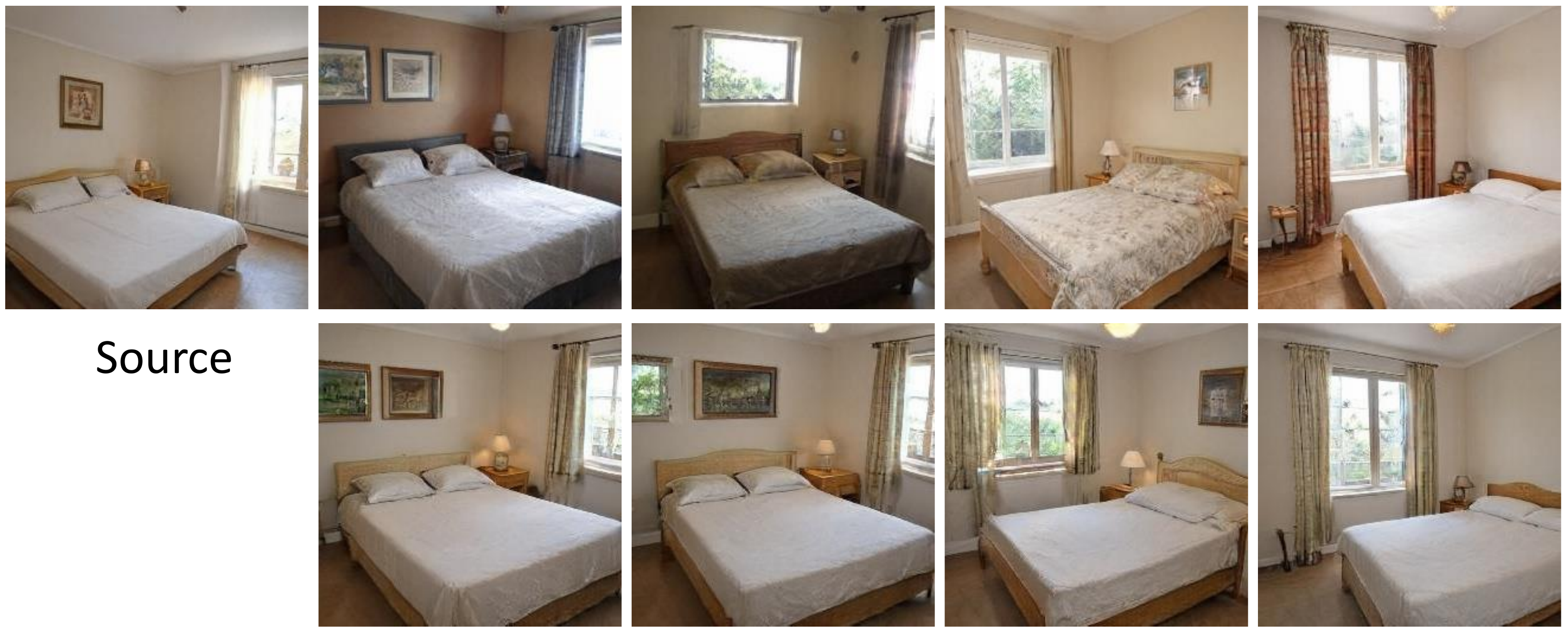}
    \caption{
        \textbf{Object rotation} with the help of clustering.
        The top-left sample indicates the source image.
        Top row represents the synthesis corresponding to four clustering centers.
        Bottom row shows the rotated beds from the source image in target poses.
    }
    \label{fig::rotating_bed}
    \vspace{-6pt}
\end{figure}

\begin{figure}[t]
    \centering
    \includegraphics[width=1.0\linewidth]{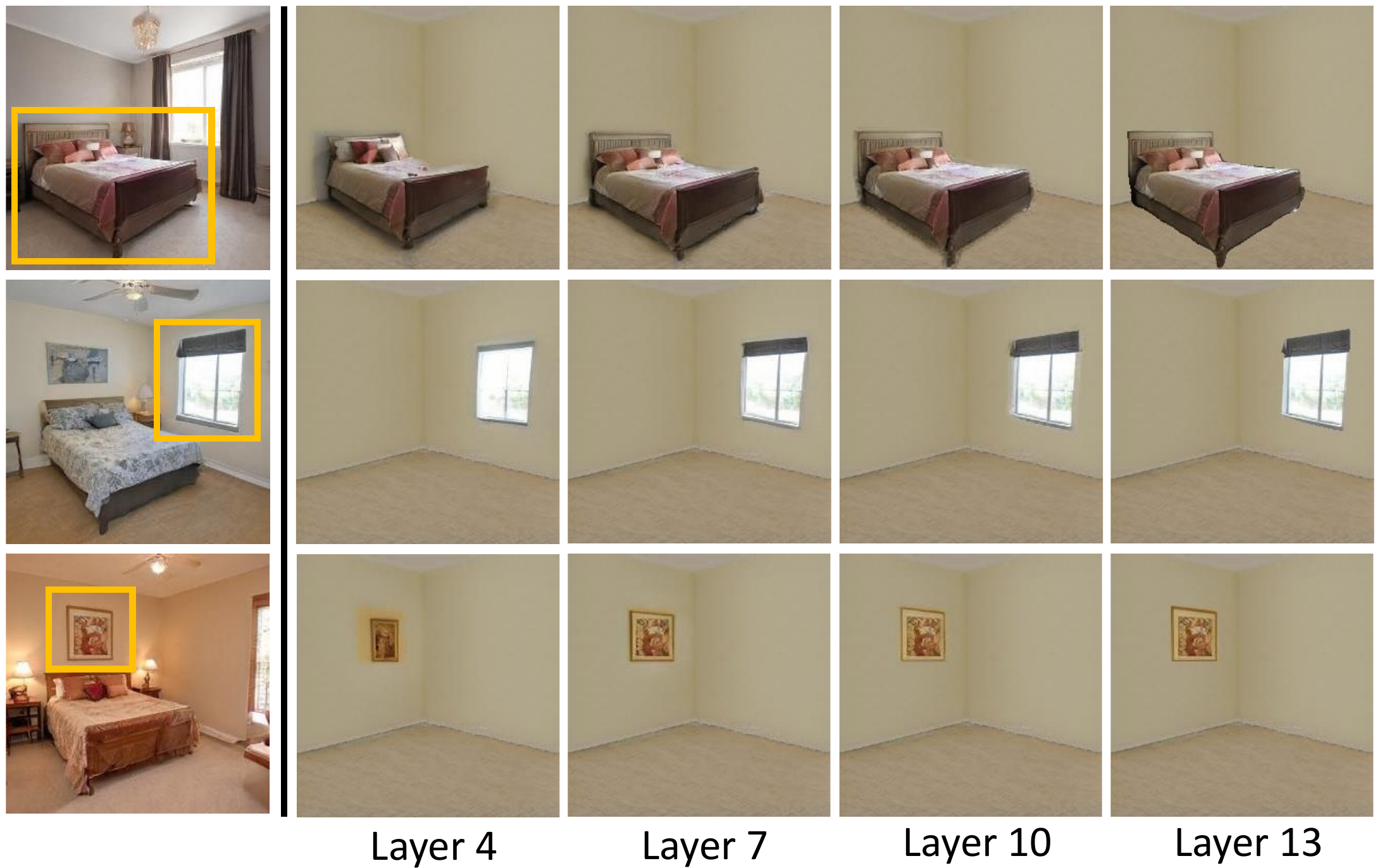}
    \caption{
        \textbf{Layer-wise analysis on object insertion}.
        We select bed, window, picture from a source image (highlighted with \textbf{\textcolor{amber}{orange}} boxes) in the first column and further insert them to the empty room at different layers.
    }
    \label{fig::paste_different_layers}
\vspace{-6pt}
\end{figure}

\vspace{-2pt}
\subsection{Inserting a New Object}
\vspace{-2pt}
Fig.~\ref{fig::paste_different_layers} presents the results of inserting the target object at the 4-th, 7-th, 10-th, 13-th layer of the generator.
The picture and window manipulation from 7-th to 13-th layers tend to give similar results, all of which can well preserve the shape and texture of the target object.
In comparison, the editing results at the 4-th layer of all these three object categories are eroded at the region boundary.
The reason is the inserted feature is blended with the background feature due to the larger receptive field of former convolutional kernels.
Consequently, we recommend to insert these objects at the 7-th layer to gain better results. 

\begin{figure}[t]
    \centering
    \includegraphics [width=1.0\linewidth]{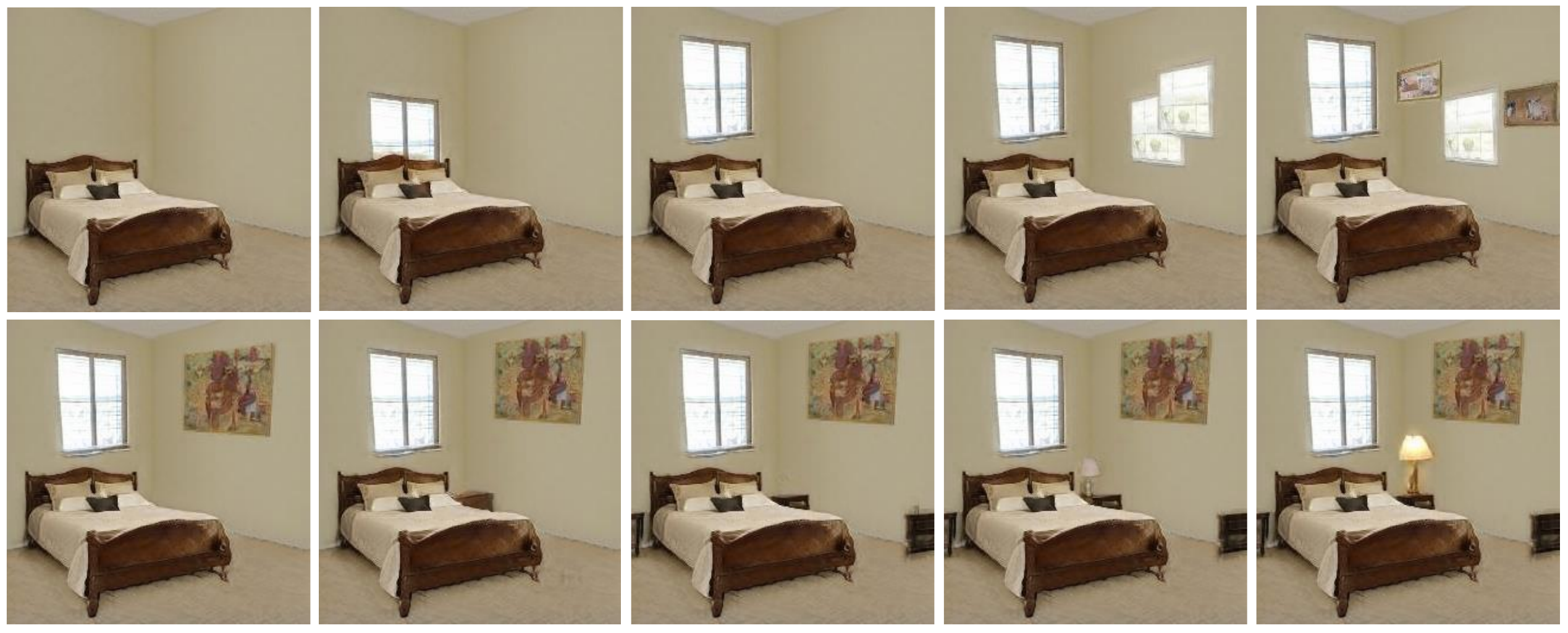}
    \caption{
        \textbf{Bedroom compositing} by progressively inserting new furniture, including windows, pictures, tables and lamps.
        LoGAN is flexible and robust to support customizing the composition.
    }
    \label{fig::compose}
\vspace{-6pt}
\end{figure}

\begin{figure}[t]
    \centering
    \includegraphics[width=1.0\linewidth]{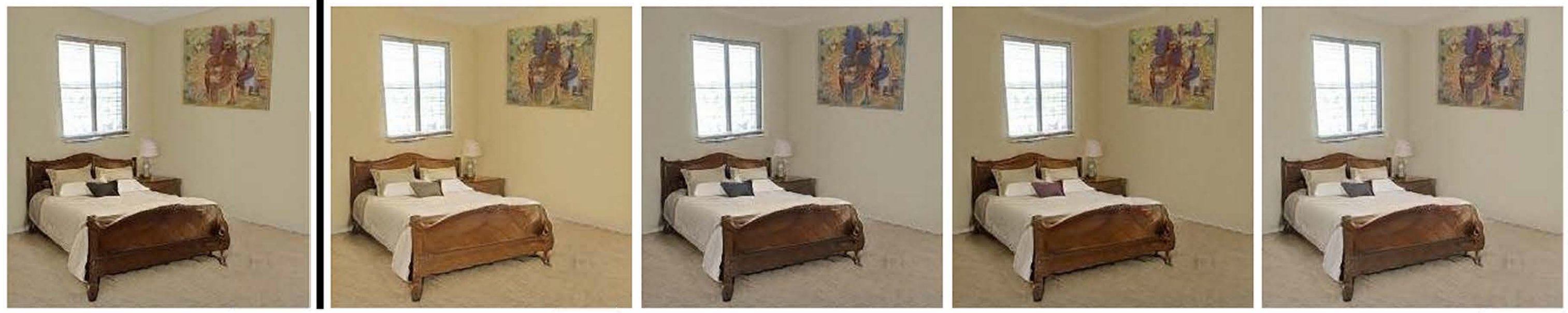}
    \caption{
        \textbf{Re-stylization} by applying new styles to the sample in the first column.
    }
    \label{fig::random_style}
    \vspace{-8pt}
\end{figure}

\vspace{-2pt}
\subsection{Bedroom Compositing}
\vspace{-2pt}
We conduct comprehensive experiments when decorating bedroom with multiple objects.
The first column of Fig.~\ref{fig::compose} acts as the base image of each row for further editing.
We first compose bed, window and picture in the generated empty bedroom, then decorate the room with diverse tables and lamps.
It's concluded that LoGAN could perfectly organize same or different objects at designated place inside the room.
The results demonstrate the flexibility and robustness of LoGAN in customizing the bedroom composition.

\vspace{-2pt}
\subsection{Changing Style}
\vspace{-2pt}
According to the formulation of StyleGAN~\cite{stylegan}, the sampled latent codes can be easily used to alter the overall image style.
In particular, we can sample numerous latent codes from the latent space and use them to guide the generation process.
Fig.~\ref{fig::random_style} visualizes some re-stylization results of the decorated room with different color schemes.

%%%% Section: Conclusion
\vspace{-2pt}
\section{Conclusion}\label{sec:conclusion}
\vspace{-2pt}
%%%%
In this work, we present LoGAN to locally control the image generation with GANs by manipulating the intermediate feature maps.
With the novel priority mask and two carefully designed modulation operators, we utilize the state-of-the-art StyleGAN2 model for photo-realistic image local editing.
This work sheds light on using well-learned GAN models to facilitate various image manipulation tasks.
We believe more works would be built on LoGAN to advance a wide range of computer graphics applications.

{\small
\bibliographystyle{ieee_fullname}
\bibliography{egbib}
}

\end{document}